\documentclass[10pt,journal,compsoc]{IEEEtran}
\usepackage{graphicx,booktabs,amsmath,microtype,array}
\usepackage[hidelinks]{hyperref}
\graphicspath{{./}}

\begin{document}
\title{On-Device Named-Entity Recognition:\\A Deployability Study of Accuracy,\\Cost, Reliability, and Confidence}
\author{Vinay~Kumar~Chaganti%
\thanks{The author is an independent researcher (e-mail: cvk.atreya@gmail.com).}%
\thanks{Portions of the evaluation code, the figure/table-generation scripts, and the manuscript prose were produced with AI-agent assistance; see the Acknowledgment.}}
\markboth{Preprint, July 2026}{Chaganti: On-Device Named-Entity Recognition: A Deployability Study}
\maketitle

\begin{abstract}
Named-entity recognition (NER) is increasingly wanted \emph{on-device} (no API, low latency, data
kept local), but the practitioner's real question is not who wins a leaderboard: it is which model is
deployable, how it is evaluated without a human-annotation budget, and whether its confidence can be
trusted. We answer these jointly. We place nine systems spanning three paradigms and 13\,M to 8\,B
parameters: a classical tagger (spaCy), bidirectional-encoder specialists (GLiNER, 166 to 460\,M), and
generative LLMs run locally (Qwen3-0.6B/1.7B/4B-Instruct, DeepSeek-R1-1.5B/8B), on three datasets of
deliberately different character, and report accuracy \emph{with} two axes the accuracy-only
literature omits: per-document latency and output validity. Because RSS-News had no gold,
we built silver gold from a cross-family LLM judge panel, then \emph{measured} its fidelity, first
against benchmark human gold and then against a full human re-validation of the corpus itself (strict
F1 0.95, an upper bound, since the human gold was seeded from silver), and show that gold provenance flips the paradigm ranking: moving from LLM-authored silver to
human gold raises every encoder and lowers every generative model. On accuracy alone a 4\,B instruct
LLM is competitive (it leads on clean newswire), so the encoder's case is deployability: it matches
or trails by a little at one-ninth to one-twenty-fourth the size, at millisecond-to-second latency,
and with \emph{zero} malformed output, while the smallest generative models emit up to 27\% invalid
output on long inputs, a pathology we show is fixed by scale, not by output budget. Finally we
characterize GLiNER's per-span confidence: it ranks correctness well (AUROC 0.76 to 0.86) but is badly
overconfident (ECE 0.24 to 0.47, roughly halved by temperature scaling); confidence thresholding gives a
small honest out-of-sample F1 gain; an all-local small$\to$large cascade gives a modest,
corpus-dependent gain over cost-matched random routing; and the confidence tracks correctness but not novelty. Every number recomputes offline from
per-span records.
\end{abstract}


\section{Introduction}
\IEEEPARstart{N}{amed}-entity recognition underlies a great deal of production text
processing (search, redaction, knowledge-base population, compliance screening), and a growing share
of it wants to run \emph{on-device}: no per-call API cost, no network round-trip, no sending sensitive
text to a third party. The practitioner who wants this faces three questions that the accuracy-centric
literature answers only in part. \emph{Which} small model should they deploy, judged not on F1 alone
but on the axes that decide deployability: latency and output reliability? \emph{How} do they
evaluate it on their own data without paying a human-annotation tax? And \emph{can} they trust the
model's confidence, to abstain or to escalate hard cases? This paper treats those three as one study.

We are explicitly not chasing a leaderboard number. We map the accuracy, cost, and reliability
\emph{frontier} across paradigms and sizes; we build an evaluation that needs \emph{no} human
annotation but whose trustworthiness we \emph{measure} rather than assume; and we characterize the
deployable recognizer's per-span confidence as a reliability tool. Our stance is deliberately honest
about what is new. That compact encoders are strong at NER is established~\cite{gliner,universalner};
that a capable LLM can match an encoder on clean text is recent~\cite{zhan}; that models are
accurate-but-overconfident~\cite{guo} and that NER confidence can be calibrated~\cite{jagannatha} are
known. Our contribution is not to rediscover these but to \emph{measure them together}, at the sub-1B
on-device operating point, with confidence intervals throughout, and to add the parts that are
genuinely underserved: output validity as a first-class metric, a \emph{measured} fidelity bound for
annotation-free NER gold, and an all-local uncertainty-gated cascade.

\textbf{Contributions.} (1)~A deployability frontier for on-device NER across three paradigms and
13\,M to 8\,B parameters on three measurably different datasets, reported on accuracy, latency, and
output validity with bootstrap confidence intervals; including the finding that the invalid-output
failure of small generative models is fixed by \emph{scale}, not output budget. (2)~An annotation-free
evaluation protocol whose fidelity we measure both on benchmarks and against a full human
re-validation of the target corpus, with the observation that LLM-authored gold systematically flips
the encoder-vs-generative ranking. (3)~A calibration, selective-prediction, and all-local cascade
study of the deployable recognizer's confidence, positioned carefully against prior NER-uncertainty
work. All artifacts reproduce offline from per-span records.

\section{Related Work}
\textbf{Small-model and generative NER.} The evidence that NER does not need a large model is
GLiNER~\cite{gliner}, which recasts NER as span-type matching in a bidirectional encoder; its ability
is itself distilled from an LLM's open-ended entity knowledge, as in
UniversalNER~\cite{universalner} and NuNER~\cite{nuner}. Zero-shot LLM NER is documented as
prompt-sensitive and error-prone~\cite{xie,ji}, and a recent assessment finds a mid-size LLM can tie a
strong encoder on CoNLL yet lose on specialized text and over-generate spans~\cite{zhan}. We place
classical, encoder, and generative systems on one measured frontier and add latency and output
validity as first-class axes.
\textbf{Annotation-free evaluation.} Using LLMs to annotate NER is
established~\cite{universalner,nuner}; panels of diverse models reduce single-judge
bias~\cite{poll}, and the alternative-annotator test formalizes justifying LLM annotators against a
human subset~\cite{alttest}. LLM-authored labels are known to shift system rankings toward LLM-style
systems in IR~\cite{clarke}, and self-preference bias is documented~\cite{selfpref}; a contemporaneous
multi-LLM NER-annotation pipeline validates with LLM-judge quality scores rather than human
gold~\cite{finerweb}. Our differentiator is a \emph{measured} strict-F1 fidelity bound (transferred
from benchmarks and then confirmed by full human re-validation), and a clean NER instance of the
provenance-driven ranking flip.
\textbf{Calibration, selective prediction, routing.} Networks are accurate but
overconfident~\cite{guo}; selective classification trades coverage for risk~\cite{geifman}. NER
specifically has span-level calibration~\cite{jagannatha}, documented
overconfidence~\cite{hashimoto}, and evidential uncertainty that beats raw confidence for
OOD~\cite{ener}. LinkNER routes uncertain mentions to a \emph{cloud} LLM~\cite{linkner}; C\textsuperscript{3} cascades small$\to$large by
calibrated confidence but for sentence-level NLU~\cite{c3}. Our cascade is the all-local,
encoder$\to$encoder, span-gated NER configuration in the gap between them.
\textbf{Difficulty and novelty.} That unseen entities drive NER difficulty is a long-standing
result~\cite{augenstein,taille}, and interpretable, attribute-bucketed evaluation~\cite{fuliu} and
model-aware difficulty measures~\cite{vinfo} are the established characterization tools; dataset-level
statistics are also used~\cite{wang}. We treat characterization as a caveated negative result, below.

\section{Study Design}
We frame the study as measurement; each choice follows from that.

\textbf{Datasets (Table~\ref{tab:data}), chosen for measurable difference.} CoNLL-2003 (clean
newswire), WNUT-17 (noisy user-generated text with emerging entities), and RSS-News, an in-house
corpus of 93 long-form articles sampled from 24 RSS feeds (technology and AI-leaning, with some
general-interest long-form). The benchmarks ship sentence-level human gold; RSS-News is document-level
($\approx$4\,k characters) and we human-validate its gold (Section~\ref{sec:anno}).

\begin{table}[!t]\centering\caption{Dataset samples (hashed, fixed).}\label{tab:data}%
\resizebox{\columnwidth}{!}{\begin{tabular}{@{}lllrrr@{}}
\toprule
Dataset & unit & gold & $n$ & entities & med.\ chars \\
\midrule
CoNLL & sentence & human & 300 & 655 & 52 \\
WNUT & sentence & human & 300 & 462 & 93 \\
RSS-News & document & human & 93 & 1760 & 4024 \\
\bottomrule
\end{tabular}
}\end{table}

\textbf{Systems, a ladder that separates paradigm from scale.} spaCy \texttt{en\_core\_web\_sm}
(classical); GLiNER small/medium/large (encoder specialist); and, run locally via Ollama with
JSON-schema-constrained decoding, Qwen3-0.6B, Qwen3-1.7B, \emph{Qwen3-4B-Instruct} (a fair mid-size
instruct model), and the reasoning models DeepSeek-R1-1.5B and R1-8B. The instruct/reasoning pair at
matched size isolates whether reasoning helps extraction. All run on the same commodity hardware
(Apple M4 Pro); none is fine-tuned.

\textbf{Metrics.} Accuracy is \texttt{nervaluate} SemEval-strict F1 (exact boundary \emph{and} type),
with 95\% confidence intervals from a 5{,}000-resample bootstrap over documents and a paired bootstrap
for head-to-head tests. We add per-document \emph{latency} and \emph{output validity}. Generative
systems decode under Ollama structured-output constraints (the JSON schema supplied as a decoding
grammar), with reasoning mode disabled and a fixed output-token budget; because decoding is
grammar-constrained, malformed tokens are impossible by construction, so an \emph{invalid} output is
one where the model fails to \emph{terminate} a schema-valid object within the budget: a truncated,
unparseable object. Every predicted span is written once to an
append-only, provenance-stamped record; every table regenerates from those records with no
re-inference, and each dataset sample is content-hashed.

\section{The Deployability Frontier}
\subsection{Accuracy: on clean text the gap is closed; encoders lead on noise}
Table~\ref{tab:frontier} reports strict F1 with CIs. On accuracy alone the fair mid-size LLM is
competitive: Qwen3-4B-Instruct leads on clean CoNLL (0.753) ahead of GLiNER-large (0.739), and leads
on RSS-News (0.684). Encoders lead where text is noisy and entities are novel:
GLiNER-large tops WNUT (0.599), with GLiNER-small (0.545) also ahead of the 4\,B model (0.506). Two
patterns matter. First, size is non-monotonic \emph{within} a paradigm: GLiNER-small (166\,M) beats
GLiNER-medium (209\,M) on WNUT by 0.093, and the tiny generative models are far below their larger
siblings. Second, reasoning is the wrong inductive bias for extraction: DeepSeek-R1-1.5B (0.398/0.264
on CoNLL/WNUT) is no better than the 0.6\,B instruct model and far below the same-size Qwen3-1.7B
(0.652/0.485). ``Encoders beat LLMs'' is thus \emph{not} the honest headline; the honest headline is
that encoders are the \emph{deployable} choice, which the next axes establish.

\begin{table*}[!t]\centering\caption{The on-device NER frontier: strict F1 [95\% CI] per dataset
(human gold throughout), with per-document latency (p50, on long RSS-News documents) and
output-invalid rate. Best per column in bold.}\label{tab:frontier}\setlength{\tabcolsep}{4pt}{\footnotesize \begin{tabular}{@{}lllcccrr@{}}
\toprule
System & Params & Mem & CoNLL & WNUT & RSS-News & lat.\,(s) & inv. \\
\midrule
spaCy (sm) & 13M & 15\,MB & 0.595\,{\tiny[0.55,0.64]} & 0.261\,{\tiny[0.22,0.30]} & 0.385\,{\tiny[0.33,0.43]} & 0.11 & 0\% \\
GLiNER-small & 166M & 0.7\,GB & 0.698\,{\tiny[0.66,0.74]} & 0.545\,{\tiny[0.50,0.58]} & 0.497\,{\tiny[0.44,0.55]} & 0.96 & 0\% \\
GLiNER-medium & 209M & 0.85\,GB & 0.706\,{\tiny[0.66,0.74]} & 0.452\,{\tiny[0.41,0.49]} & 0.479\,{\tiny[0.43,0.53]} & 1.68 & 0\% \\
GLiNER-large & 460M & 1.8\,GB & 0.739\,{\tiny[0.70,0.78]} & \textbf{0.599}\,{\tiny[0.55,0.64]} & 0.603\,{\tiny[0.56,0.64]} & 4.02 & 0\% \\
Qwen3-0.6B & 596M & 1.5\,GB & 0.408\,{\tiny[0.37,0.45]} & 0.236\,{\tiny[0.20,0.28]} & 0.243\,{\tiny[0.18,0.30]} & 2.73 & 27\% \\
Qwen3-1.7B & 1.7B & 2.2\,GB & 0.652\,{\tiny[0.61,0.69]} & 0.485\,{\tiny[0.44,0.53]} & 0.532\,{\tiny[0.48,0.58]} & 3.56 & 8\% \\
Qwen3-4B-Instruct & 4B & 4.3\,GB & \textbf{0.753}\,{\tiny[0.71,0.79]} & 0.506\,{\tiny[0.47,0.55]} & \textbf{0.684}\,{\tiny[0.63,0.73]} & 5.10 & 0\% \\
DeepSeek-R1-1.5B & 1.5B & 1.9\,GB & 0.398\,{\tiny[0.34,0.45]} & 0.264\,{\tiny[0.22,0.30]} & 0.217\,{\tiny[0.16,0.28]} & 5.50 & 14\% \\
DeepSeek-R1-8B & 8B & 5.2\,GB & 0.712\,{\tiny[0.67,0.75]} & 0.461\,{\tiny[0.41,0.51]} & 0.578\,{\tiny[0.52,0.63]} & 28.33 & 5\% \\
\bottomrule
\end{tabular}}
\end{table*}

\begin{figure}[!t]\centering\includegraphics[width=\columnwidth]{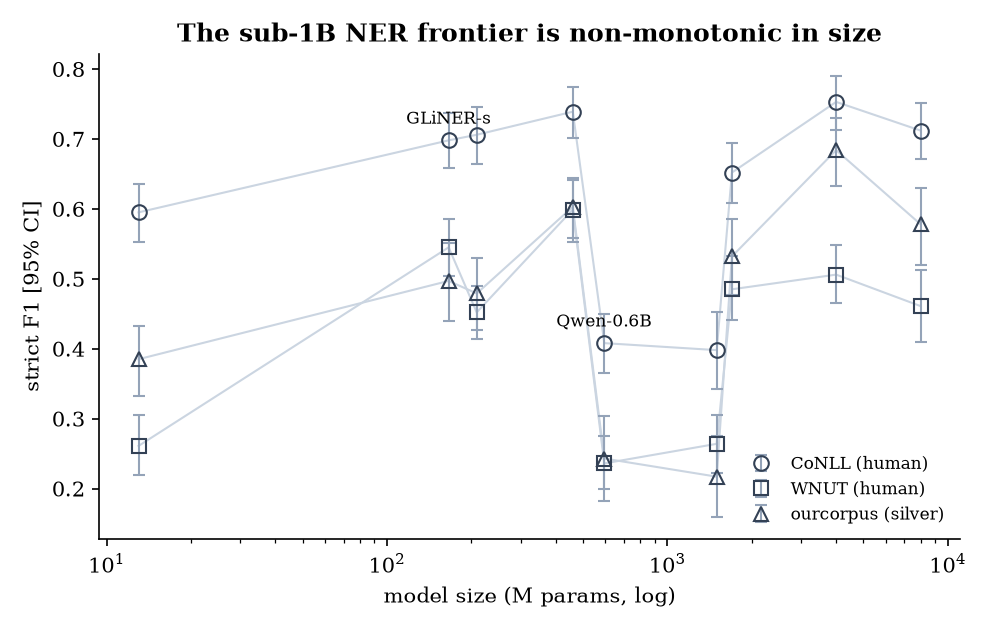}
\caption{Strict F1 (95\% CI) versus model size. The useful frontier is non-monotonic in
size.}\label{fig:frontier}\end{figure}

\subsection{Reliability and latency: where the generative arm is disqualified}
\begin{figure}[!t]\centering\includegraphics[width=\columnwidth]{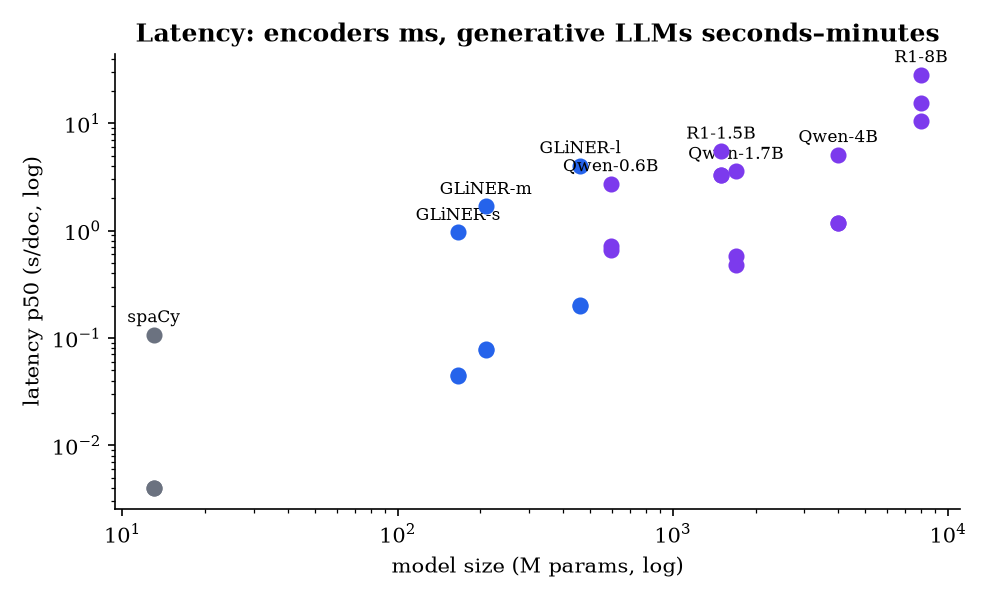}
\caption{Per-document latency versus size (log-log): encoders in milliseconds to a few seconds,
generative LLMs in seconds to minutes.}\label{fig:latency}\end{figure}

Output \emph{validity} is a long-document, size-dependent failure of the generative arm
(Table~\ref{tab:valid}). On the $\approx$4\,k-character documents Qwen3-0.6B emits 27\% invalid output
and DeepSeek-R1-1.5B 14\%, while the encoders and spaCy \emph{never} fail. Crucially the failure is \emph{non-termination}, not malformed decoding, and more
budget does not fix it: raising the token budget from 1{,}200 to 4{,}000 barely
moves the 0.6\,B model (27\%$\to$25\% invalid) while its worst-case latency explodes to 97\,s, whereas
Qwen3-4B-Instruct is at 0\% invalid under \emph{either} budget. Validity is fixed by \emph{scale},
not budget: a small generative model fails to reliably \emph{terminate} valid structure on long
inputs even under schema-constrained decoding, exactly the regime an on-device deployment lives in.
Latency compounds this (Fig.~\ref{fig:latency}): encoders run in 0.1 to 4\,s, the small LLMs in seconds,
and R1-8B at 28\,s per document. On-device, the 8\,B model is disqualified, not merely expensive.

\begin{table}[!t]\centering\caption{Output-validity decomposition on long RSS-News documents:
invalid-output rate, latency (s), and strict F1 at two output-token budgets.}\label{tab:valid}%
\resizebox{\columnwidth}{!}{\begin{tabular}{@{}lrrrrr@{}}
\toprule
Model & $n_{\mathrm{pred}}$ & invalid & lat.\ p50 & lat.\ max & F1 \\
\midrule
Qwen3-0.6B & 1200 & 26.9\% & 2.7 & 16 & 0.243 \\
Qwen3-0.6B & 4000 & 24.7\% & 2.1 & 97 & 0.234 \\
Qwen3-1.7B & 1200 & 7.5\% & 3.6 & 23 & 0.532 \\
Qwen3-1.7B & 4000 & 7.5\% & 2.3 & 49 & 0.532 \\
Qwen3-4B-Instruct & 1200 & 0.0\% & 5.1 & 15 & 0.684 \\
Qwen3-4B-Instruct & 4000 & 0.0\% & 5.0 & 15 & 0.684 \\
\bottomrule
\end{tabular}
}\end{table}

\subsection{Per-type and significance}
The paradigms fail differently by type (Table~\ref{tab:pertype}): PERSON is easiest; PRODUCT is
hardest, and the classical tagger scores near zero on it because its fixed schema has no slot for
it, a structural, not statistical, limit. Paired bootstrap tests (Table~\ref{tab:sig}) confirm the
GLiNER-small\,$>$\,GLiNER-medium WNUT gap and the tiny-LLM deficits are real.

\begin{table*}[!t]\centering\caption{Per-type strict F1 on WNUT and CoNLL.}\label{tab:pertype}\setlength{\tabcolsep}{4pt}{\footnotesize \begin{tabular}{@{}lcccc|ccc@{}}
\toprule
& \multicolumn{4}{c}{WNUT} & \multicolumn{3}{c}{CoNLL} \\
\cmidrule(lr){2-5}\cmidrule(lr){6-8}
System & PER & ORG & LOC & PROD & PER & ORG & LOC \\
\midrule
spaCy (sm) & 0.295 & 0.211 & 0.406 & 0.000 & 0.662 & 0.360 & 0.761 \\
GLiNER-small & 0.694 & 0.385 & 0.577 & 0.421 & 0.890 & 0.608 & 0.627 \\
GLiNER-medium & 0.645 & 0.249 & 0.548 & 0.308 & 0.879 & 0.541 & 0.706 \\
GLiNER-large & 0.714 & 0.483 & 0.654 & 0.422 & 0.893 & 0.673 & 0.660 \\
Qwen3-0.6B & 0.288 & 0.289 & 0.105 & 0.080 & 0.555 & 0.432 & 0.124 \\
Qwen3-1.7B & 0.622 & 0.341 & 0.485 & 0.278 & 0.821 & 0.483 & 0.641 \\
Qwen3-4B-Instruct & 0.696 & 0.489 & 0.579 & 0.224 & 0.909 & 0.622 & 0.745 \\
DeepSeek-R1-1.5B & 0.308 & 0.243 & 0.253 & 0.186 & 0.619 & 0.255 & 0.328 \\
DeepSeek-R1-8B & 0.547 & 0.388 & 0.525 & 0.316 & 0.908 & 0.611 & 0.628 \\
\bottomrule
\end{tabular}}
\end{table*}
\begin{table*}[!t]\centering\caption{Paired significance ($\Delta$ strict F1 $=$ A$-$B, 95\% CI).}\label{tab:sig}\setlength{\tabcolsep}{4pt}{\footnotesize \begin{tabular}{@{}lccc@{}}
\toprule
Comparison ($\Delta$ strict F1) & CoNLL & WNUT & RSS-News \\
\midrule
GLiNER-small $-$ Qwen3-0.6B & $+0.291$\,{\tiny[+0.24,+0.35]}\,sig & $+0.309$\,{\tiny[+0.26,+0.36]}\,sig & $+0.254$\,{\tiny[+0.18,+0.33]}\,sig \\
GLiNER-small $-$ GLiNER-medium & $-0.007$\,{\tiny[-0.04,+0.03]}\,n.s. & $+0.093$\,{\tiny[+0.06,+0.12]}\,sig & $+0.017$\,{\tiny[-0.03,+0.06]}\,n.s. \\
GLiNER-small $-$ Qwen3-1.7B & $+0.047$\,{\tiny[+0.00,+0.10]}\,sig & $+0.060$\,{\tiny[+0.01,+0.11]}\,sig & $-0.036$\,{\tiny[-0.10,+0.03]}\,n.s. \\
GLiNER-medium $-$ spaCy & $+0.110$\,{\tiny[+0.07,+0.15]}\,sig & $+0.191$\,{\tiny[+0.14,+0.24]}\,sig & $+0.094$\,{\tiny[+0.05,+0.14]}\,sig \\
\bottomrule
\end{tabular}}
\end{table*}

\section{Annotation-Free Evaluation and Its Fidelity}\label{sec:anno}
\subsection{A measured fidelity bound}
For the unlabeled corpus we generate silver gold from a three-model cross-family judge panel
(Gemini-Flash-Lite, Nemotron-3-Ultra-550B, and Mistral-Small-119B), taking the $\ge$2-of-3 span consensus, a
different model family from every system under test, so the gold is not a system grading itself. This panel is a cloud, large-model evaluation \emph{oracle}, not part of the deployed system: using it to label a corpus one would otherwise keep on-device is a one-time evaluation cost and a genuine tension we do not hide: a fully local judge panel is future work. We do
not assume this is trustworthy; we \emph{measure} it (Table~\ref{tab:fid}). Against benchmark human
gold the panel reaches strict F1 0.834 (CoNLL) and 0.749 (WNUT). We then hand-validated all 93
RSS-News documents: the panel's silver gold reaches strict F1 0.952 (95\% CI [0.937, 0.965]) against
that human gold, with precision 0.987 and recall 0.919. This 0.952 is an \emph{upper bound}: the human gold was seeded from the silver spans (annotators edited panel output), so measured acceptance is optimistically biased. The panel is thus high-precision and its only
material weakness is recall: it \emph{misses} entities rather than inventing them. The human pass
accepted 98.7\% of silver spans unchanged, added 143 misses, deleted only 22 (17 of them the fuzzy
PRODUCT type), and re-typed none.

\begin{table}[!t]\centering\caption{Silver-gold fidelity: judge-panel consensus versus human gold.
$^\dagger$RSS-News gold is human-validated from a silver seed.}\label{tab:fid}\begin{tabular}{@{}llccc@{}}
\toprule
Dataset & scoring gold & panel strict F1 & P & R \\
\midrule
CoNLL & human & 0.834 & n/a & n/a \\
WNUT & human & 0.749 & n/a & n/a \\
RSS-News & human$^\dagger$ & \textbf{0.952}\,{\tiny[0.94,0.96]} & 0.987 & 0.919 \\
\bottomrule
\end{tabular}
\end{table}

\subsection{Gold provenance flips the ranking}
The fidelity bound is not academic (Table~\ref{tab:prov}): moving from LLM-authored silver to human
gold \emph{raises every encoder} (up to $+0.037$) and \emph{lowers every generative model} (down to
$-0.029$), a clean split by paradigm. Because the human gold was seeded from silver, this design is \emph{biased against} observing a flip (annotators saw and largely kept the panel's spans), so the flip is a conservative estimate of the provenance effect. The mechanism is that silver, authored by LLMs, both flattered
LLM-style output and under-credited entities the encoders correctly found but the panel missed
(scored as false positives against silver, vindicated by human gold). The visible consequence is a
rank change: GLiNER-large overtakes R1-8B on human gold. This is the NER instance of a
provenance-driven ranking flip previously shown in IR~\cite{clarke}, and it is why we anchor all
accuracy claims on human gold.

\begin{table}[!t]\centering\caption{Provenance effect on RSS-News: strict F1 under silver
vs.\ human gold ($\Delta$ bold when $|\Delta|\ge0.02$). enc/gen/cls = encoder/generative/classical.}%
\label{tab:prov}\begin{tabular}{@{}lccc@{}}
\toprule
System & F1 vs silver & F1 vs human & $\Delta$ \\
\midrule
spaCy (cls) & 0.365 & 0.385 & +0.020 \\
GLiNER-small (enc) & 0.482 & 0.517 & \textbf{+0.036} \\
GLiNER-medium (enc) & 0.468 & 0.500 & \textbf{+0.032} \\
GLiNER-large (enc) & 0.582 & 0.619 & \textbf{+0.037} \\
Qwen3-0.6B (gen) & 0.249 & 0.243 & -0.006 \\
Qwen3-1.7B (gen) & 0.550 & 0.532 & -0.017 \\
Qwen3-4B-Instruct (gen) & 0.713 & 0.684 & \textbf{-0.029} \\
DeepSeek-R1-1.5B (gen) & 0.229 & 0.217 & -0.011 \\
DeepSeek-R1-8B (gen) & 0.608 & 0.578 & \textbf{-0.029} \\
\bottomrule
\end{tabular}
\end{table}

\section{Confidence: Calibration, Selective Prediction, Routing}
The three GLiNER encoders emit a native per-span confidence; the classical and generative systems do
not, so this section concerns the deployable recognizer family.

\subsection{Ranks correctness, but miscalibrated}
\begin{figure}[!t]\centering\includegraphics[width=\columnwidth]{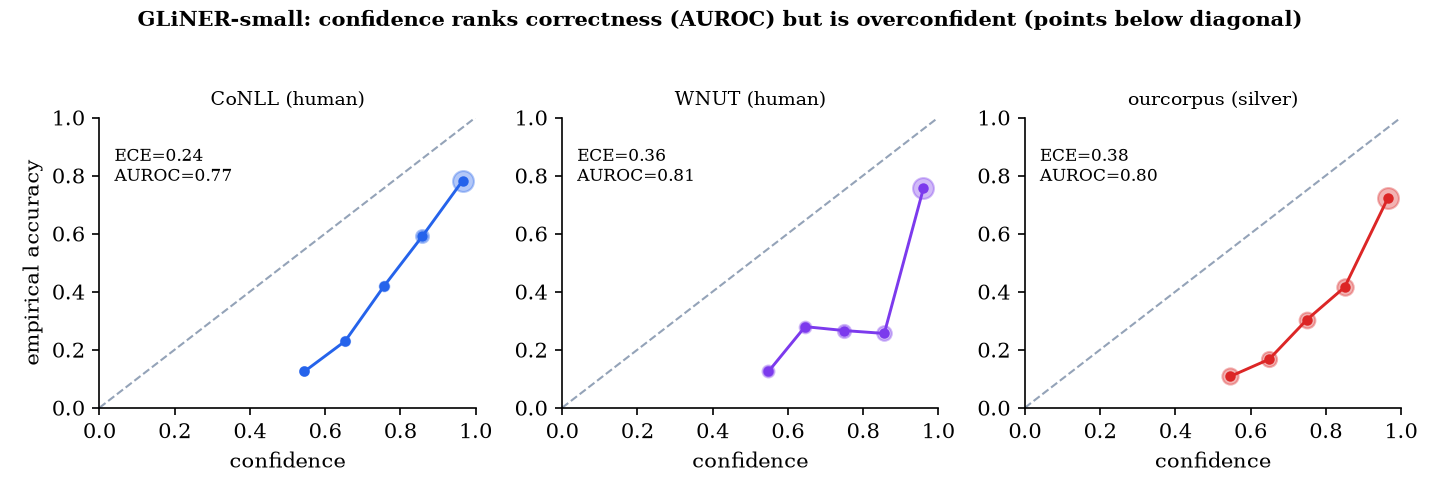}
\caption{Reliability diagrams (GLiNER-small); points below the diagonal indicate overconfidence, while
accuracy rising with confidence indicates usable ranking. Rendered from measured data; see the
Acknowledgment.}\label{fig:rel}\end{figure}

The confidence separates correct from incorrect spans everywhere (AUROC 0.76 to 0.86,
Table~\ref{tab:calib}, Fig.~\ref{fig:rel}) but is badly overconfident (ECE 0.24 to 0.47; every
reliability point below the diagonal). Temperature scaling (fit by NLL on the inverted GLiNER
confidence) roughly halves ECE (e.g., 0.45$\to$0.28 on RSS-News), with fitted
temperatures far above one, confirming systematic overconfidence. Span-level NER calibration is not
new~\cite{jagannatha,guo}; the contribution here is characterizing GLiNER's native confidence and
showing a post-hoc fix suffices. The operational reading: do not read the score as a probability, but
do trust its ordering.

\begin{table*}[!t]\centering\caption{Confidence separability (AUROC [95\% CI]) and calibration
(ECE\,$\to$\,ECE after temperature scaling $T$).}\label{tab:calib}\setlength{\tabcolsep}{4pt}{\footnotesize \begin{tabular}{@{}lcccccc@{}}
\toprule
& \multicolumn{2}{c}{CoNLL} & \multicolumn{2}{c}{WNUT} & \multicolumn{2}{c}{RSS-News} \\
\cmidrule(lr){2-3}\cmidrule(lr){4-5}\cmidrule(lr){6-7}
System & AUROC\,[CI] & ECE$\to T$ & AUROC\,[CI] & ECE$\to T$ & AUROC\,[CI] & ECE$\to T$ \\
\midrule
GLiNER-small & 0.77\,{\tiny[0.73,0.81]} & 0.24$\to$0.11 & 0.81\,{\tiny[0.77,0.85]} & 0.36$\to$0.19 & 0.80\,{\tiny[0.77,0.83]} & 0.38$\to$0.23 \\
GLiNER-medium & 0.76\,{\tiny[0.72,0.80]} & 0.27$\to$0.10 & 0.85\,{\tiny[0.82,0.87]} & 0.47$\to$0.18 & 0.86\,{\tiny[0.84,0.88]} & 0.45$\to$0.28 \\
GLiNER-large & 0.77\,{\tiny[0.72,0.81]} & 0.25$\to$0.09 & 0.78\,{\tiny[0.73,0.82]} & 0.35$\to$0.13 & 0.81\,{\tiny[0.78,0.84]} & 0.29$\to$0.17 \\
\bottomrule
\end{tabular}}
\end{table*}

\subsection{Selective prediction, honestly measured}
\begin{figure}[!t]\centering\includegraphics[width=\columnwidth]{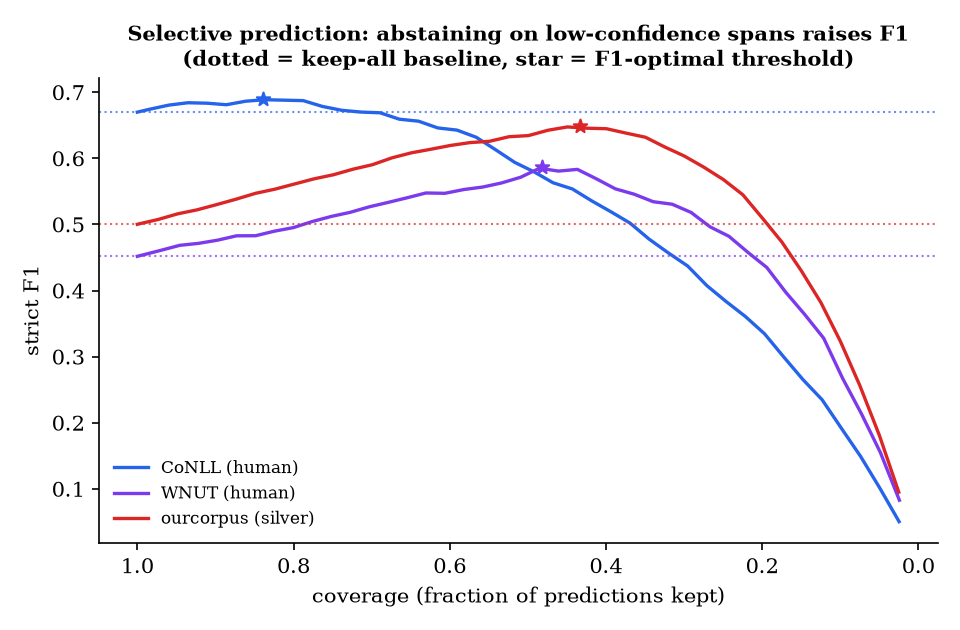}
\caption{Strict F1 versus coverage as the confidence threshold sweeps (GLiNER-medium); abstaining on
low-confidence spans raises F1, most on the harder datasets.}\label{fig:sel}\end{figure}

Because the ranking is good, thresholding the confidence to abstain on low-confidence spans is useful
(Table~\ref{tab:sel}, Fig.~\ref{fig:sel}). We select the threshold on a validation split and apply it
to a disjoint test split (averaged over 40 random splits), so we report a \emph{transferred},
out-of-sample gain rather than a test-tuned upper bound. Most gains are modest ($+0.01$ to $+0.06$
strict F1), but for the badly-miscalibrated GLiNER-medium the gain is large and survives transfer
($+0.133$ WNUT, $+0.143$ RSS-News): selective prediction helps most exactly where calibration is
worst. We report threshold-free AURC as the primary object. This differs from evidential NER
uncertainty~\cite{ener}, which learns an uncertainty signal to flag entities rather than abstaining
under a validation-transferred threshold.

\begin{table*}[!t]\centering\caption{Selective prediction: keep-all strict F1 $\to$ F1 at a threshold
chosen on validation and transferred to test (mean\,$\pm$\,sd over 40 splits).}\label{tab:sel}\setlength{\tabcolsep}{4pt}{\footnotesize \begin{tabular}{@{}lccc@{}}
\toprule
System (keep-all\,$\to$\,transferred F1) & CoNLL & WNUT & RSS-News \\
\midrule
GLiNER-small & 0.685\,$\to$\,0.701 (+0.016\,$\pm$\,0.010) & 0.545\,$\to$\,0.568 (+0.023\,$\pm$\,0.013) & 0.517\,$\to$\,0.574 (+0.056\,$\pm$\,0.016) \\
GLiNER-medium & 0.670\,$\to$\,0.681 (+0.011\,$\pm$\,0.008) & 0.452\,$\to$\,0.585 (+0.133\,$\pm$\,0.014) & 0.500\,$\to$\,0.643 (+0.143\,$\pm$\,0.019) \\
GLiNER-large & 0.720\,$\to$\,0.733 (+0.013\,$\pm$\,0.008) & 0.601\,$\to$\,0.625 (+0.024\,$\pm$\,0.015) & 0.619\,$\to$\,0.643 (+0.024\,$\pm$\,0.016) \\
\bottomrule
\end{tabular}}
\end{table*}

\subsection{An all-local confidence-gated cascade}
\begin{figure}[!t]\centering\includegraphics[width=\columnwidth]{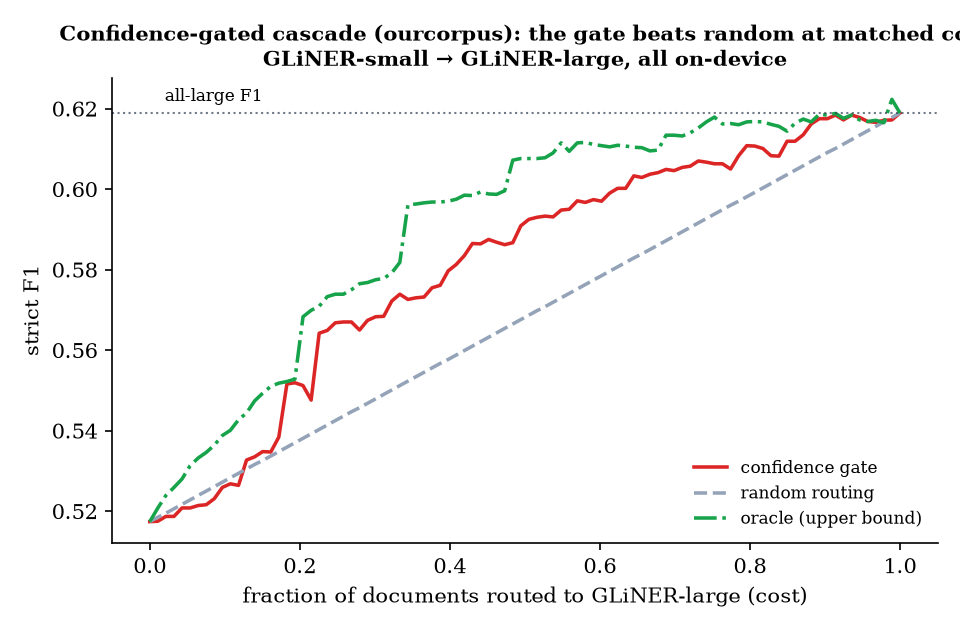}
\caption{Strict F1 versus fraction of documents routed to GLiNER-large (RSS-News). The gate
is ahead of random, but the margin is significant only on this corpus; the whole cascade runs
on-device.}\label{fig:route}\end{figure}

The same ranking drives a cascade that never leaves the device (Table~\ref{tab:route},
Fig.~\ref{fig:route}): GLiNER-small is the base, and the most-uncertain documents are routed to
GLiNER-large, the local stronger tier, not a cloud LLM as in LinkNER~\cite{linkner}. Against a
cost-matched \emph{random} routing baseline the gate is ahead at every budget, but the margin is
small and its 95\% bootstrap CI excludes zero \emph{only} on RSS-News ($+0.021$
[$+0.001,+0.044$] at 50\% routed); on the benchmarks the gate$-$random gap is within noise ($+0.016$
[$-0.003,+0.035$] CoNLL, $+0.004$ [$-0.014,+0.024$] WNUT). Confidence-gated routing thus gives a
modest, corpus-dependent improvement over spending the same compute at random, bounded above by an
oracle. The mechanism (uncertainty routing) is prior art~\cite{linkner,c3}; the contribution
is the specific all-local encoder$\to$encoder configuration.

\begin{table}[!t]\centering\caption{All-local cascade (GLiNER-small\,$\to$\,GLiNER-large): endpoint
F1, gate/random F1 at 25\% and 50\% routed, and $f_{95}$ (routed fraction reaching 95\% of the full
gain).}\label{tab:route}\resizebox{\columnwidth}{!}{\setlength{\tabcolsep}{4pt}\resizebox{\columnwidth}{!}{\begin{tabular}{@{}lccccc@{}}
\toprule
Dataset & small & large & gate/rand@50 & $\Delta$@50\,[CI] & $f_{95}$ \\
\midrule
CoNLL & 0.685 & 0.720 & 0.718/0.703 & +0.016\,{\tiny[-0.00,+0.04]} & 0.42 \\
WNUT & 0.545 & 0.601 & 0.575/0.573 & +0.004\,{\tiny[-0.01,+0.02]} & 0.60 \\
RSS-News & 0.517 & 0.619 & 0.592/0.569 & +0.021\,{\tiny[+0.00,+0.04]} & 0.88 \\
\bottomrule
\end{tabular}}
}\end{table}

\subsection{Confidence tracks correctness, not novelty}
Per document, GLiNER-small's mean confidence correlates with strict F1 (Spearman $\rho=+0.30$ CoNLL,
$+0.46$ WNUT, both $p<10^{-4}$) but not with the unseen-entity ratio ($+0.07$, $-0.05$, both n.s.,
Table~\ref{tab:nov}). The confidence reflects internal certainty about its own decision, not the
out-of-distribution nature of the input, consistent with evidential-uncertainty
findings~\cite{ener}. It is a sound basis for abstention and routing but is \emph{not} an OOD detector.

\begin{table}[!t]\centering\caption{Spearman correlation of GLiNER-small per-document mean confidence
with document strict F1 and with unseen-entity ratio. $^{*}$: $p<0.05$.}\label{tab:nov}\begin{tabular}{@{}lccc@{}}
\toprule
Dataset & $n$ & $\rho$(conf, unseen) & $\rho$(conf, F1) \\
\midrule
CoNLL & 299 & $+0.070$ & $+0.299$$^{*}$ \\
WNUT & 291 & $-0.051$ & $+0.461$$^{*}$ \\
\bottomrule
\end{tabular}
\end{table}

\section{What Makes It Hard: A Characterization Caveat}
We had hoped measurable dataset characteristics would predict difficulty. They do not, at the
resolution one would want, and the effect that survives is not new. Within each dataset, per-document
correlations between static characteristics (entity density, length, count, even unseen ratio) and F1
are weak and dataset-specific ($|\rho|<0.17$, mostly not significant). What tracks the cross-dataset
ordering is entity \emph{novelty}: CoNLL's test entities are 46\% unseen in training, WNUT's 98\%
(Table~\ref{tab:char}). But this is a known result~\cite{augenstein}, unseen-entity ratio is itself
one of the dataset-level statistics~\cite{wang}, and we did not use the model-aware per-instance
difficulty measures designed for this~\cite{vinfo}. We therefore report this as a caveated negative
result: off-the-shelf \emph{per-document} static metrics do not predict where an on-device NER model
fails; only the coarse between-dataset novelty property, long known, tracks difficulty.

\begin{table}[!t]\centering\caption{Dataset characterization~\cite{wang} plus unseen-entity ratio and
model differentiation.}\label{tab:char}\resizebox{\columnwidth}{!}{\begin{tabular}{@{}lccc@{}}
\toprule
Metric & CoNLL & WNUT & RSS-News \\
\midrule
entity density & 0.128 & 0.076 & 0.042 \\
entity ambiguity (EnAmb) & 0.018 & 0.002 & 0.007 \\
type imbalance & 0.032 & 0.123 & 0.104 \\
null-doc rate & 0.000 & 0.000 & 0.054 \\
\textbf{unseen-entity ratio} & \textbf{0.458} & \textbf{0.978} & n/a \\
model differentiation & 0.129 & 0.127 & 0.150 \\
\bottomrule
\end{tabular}
}\end{table}

\section{Discussion}
\textbf{Deploy the encoder for the deployment, not the leaderboard.} A fair 4\,B instruct LLM can
match or beat the encoders on clean and long-form text, so the encoder's case is not raw accuracy: it
is 0\% malformed output, millisecond-to-second latency, and one-ninth to one-twenty-fourth the
parameters. GLiNER-large is the accuracy-leaning on-device choice; GLiNER-small the efficiency default.
\textbf{Rank, do not read, the confidence.} Used as a probability it is wrong (ECE up to 0.47); used
to order spans it is a free, reliable tool for abstention and an all-local cascade. Temperature
scaling is the fix where a probability is needed.
\textbf{Evaluate without annotation, but measure the gold.} A cross-family panel gives high-precision
silver gold; report its fidelity, and remember that LLM-authored gold flatters LLM-style systems.

\section{Limitations}
RSS-News is a single English news domain; its human gold, though validated span-by-span, was
seeded from silver, which could bias reviewers toward acceptance (the near-zero deletion rate is
consistent with genuine high precision, but we note the risk). The confidence analysis covers the
GLiNER family only. The characterization result rests on two benchmark points for the novelty axis and
does not use model-aware difficulty measures. No system is fine-tuned; a distilled on-device student is
future work.

\section{Conclusion}
For sub-1B on-device NER the deployable choice is a compact bidirectional encoder, not because
generative models cannot do NER (a 4\,B instruct model is competitive on accuracy) but because the
encoder delivers that accuracy with zero malformed output, low latency, and a fraction of the
parameters, and its invalid-output advantage over small generative models is a scale property, not a
tuning artifact. Such systems can be evaluated without human annotation via a judge-panel silver gold
whose fidelity we measure at 0.95 on the target corpus (an upper bound, as the human gold was silver-seeded), bearing in mind that LLM-authored gold flips
the paradigm ranking. And the deployable recognizer's confidence, though miscalibrated, ranks
correctness well enough to drive selective prediction and an all-local cascade. A distilled on-device
NER student, and the question of whether a fully local teacher can match a cloud one, follows.

\section*{Acknowledgment}
The evaluation harness, the figure- and table-generation scripts (which render measured data; no
result is AI-generated), and portions of the manuscript prose were produced with the assistance of AI
coding and writing agents. All experimental design, analysis decisions, and interpretation are the
author's.


\begin{thebibliography}{99}
\bibitem{gliner} U.~Zaratiana \emph{et al.}, ``GLiNER: Generalist model for NER using bidirectional transformer,'' in \emph{Proc. NAACL}, 2024.
\bibitem{universalner} W.~Zhou \emph{et al.}, ``UniversalNER: Targeted distillation from LLMs for open NER,'' in \emph{Proc. ICLR}, 2024.
\bibitem{nuner} S.~Bogdanov \emph{et al.}, ``NuNER: Entity recognition encoder pre-training via LLM-annotated data,'' in \emph{Proc. EMNLP}, 2024.
\bibitem{xie} X.~Xie \emph{et al.}, ``Empirical study of zero-shot NER with ChatGPT,'' in \emph{Proc. EMNLP}, 2023.
\bibitem{ji} B.~Ji \emph{et al.}, ``Evaluating large language models on named entity recognition,'' \emph{IEEE Trans. Neural Netw. Learn. Syst.}, 2026.
\bibitem{zhan} Q.~Zhan, Y.~Wang, and H.~Huang, ``Assessment of generative NER in the era of large language models,'' arXiv:2601.17898, 2026.
\bibitem{poll} P.~Verga \emph{et al.}, ``Replacing judges with juries: Evaluating LLM generations with a panel of diverse models,'' arXiv:2404.18796, 2024.
\bibitem{alttest} N.~Calderon, R.~Reichart, and R.~Dror, ``The alternative annotator test for LLM-as-a-judge,'' arXiv:2501.10970, 2025.
\bibitem{clarke} C.~L.~A.~Clarke and L.~Dietz, ``LLM-based relevance assessment still cannot replace human relevance assessment,'' arXiv:2412.17156, 2024.
\bibitem{selfpref} K.~Wataoka, T.~Takahashi, and R.~Ri, ``Self-preference bias in LLM-as-a-judge,'' arXiv:2410.21819, 2024.
\bibitem{finerweb} J.~Golde, F.~Haller, and A.~Akbik, ``FiNERweb: Datasets and artifacts for scalable multilingual named entity recognition,'' in \emph{Findings of EACL}, 2026, arXiv:2512.13884.
\bibitem{guo} C.~Guo, G.~Pleiss, Y.~Sun, and K.~Q.~Weinberger, ``On calibration of modern neural networks,'' in \emph{Proc. ICML}, 2017.
\bibitem{geifman} Y.~Geifman and R.~El-Yaniv, ``Selective classification for deep neural networks,'' in \emph{Proc. NeurIPS}, 2017.
\bibitem{jagannatha} A.~Jagannatha and H.~Yu, ``Calibrating structured output predictors for natural language processing,'' in \emph{Proc. ACL}, 2020.
\bibitem{hashimoto} W.~Hashimoto, H.~Kamigaito, and T.~Watanabe, ``Are data augmentation methods in named entity recognition applicable for uncertainty estimation?,'' in \emph{Proc. EMNLP}, 2024.
\bibitem{ener} Z.~Zhang \emph{et al.}, ``E-NER: Evidential deep learning for trustworthy named entity recognition,'' in \emph{Findings of ACL}, 2023.
\bibitem{linkner} Z.~Zhang \emph{et al.}, ``LinkNER: Linking local NER models to large language models using uncertainty,'' in \emph{Proc. ACM Web Conf. (WWW)}, 2024, arXiv:2402.10573.
\bibitem{c3} T.~Lu \emph{et al.}, ``C\textsuperscript{3}: Confidence calibration model cascade for inference-efficient cross-lingual NLU,'' arXiv:2402.15991, 2024.
\bibitem{augenstein} I.~Augenstein, L.~Derczynski, and K.~Bontcheva, ``Generalisation in named entity recognition: A quantitative analysis,'' \emph{Computer Speech and Language}, vol.~44, 2017.
\bibitem{taille} B.~Taill\'e, V.~Guigue, and P.~Gallinari, ``Contextualized embeddings in named-entity recognition: An empirical study on generalization,'' in \emph{Proc. ECIR}, 2020.
\bibitem{fuliu} J.~Fu, P.~Liu, and G.~Neubig, ``Interpretable multi-dataset evaluation for named entity recognition,'' in \emph{Proc. EMNLP}, 2020.
\bibitem{vinfo} K.~Ethayarajh, Y.~Choi, and S.~Swayamdipta, ``Understanding dataset difficulty with $\mathcal{V}$-usable information,'' in \emph{Proc. ICML}, 2022.
\bibitem{wang} C.~Wang, Q.~Dong, X.~Wang, and Z.~Sui, ``Statistical dataset evaluation: A case study on NER,'' \emph{Natural Language Processing}, 2025.
\end{thebibliography}
\end{document}